\documentclass{article}

\PassOptionsToPackage{numbers, compress}{natbib}

\usepackage{PRIMEarxiv}
\usepackage{color}
\usepackage{natbib}
\usepackage{wrapfig}

\usepackage{amssymb}
\usepackage[utf8]{inputenc}
\usepackage[T1]{fontenc}
\usepackage{hyperref}
\usepackage{url}
\usepackage{booktabs}
\usepackage{amsfonts}
\usepackage{nicefrac}
\usepackage{microtype}
\usepackage{array}
\usepackage{multicol}
\usepackage{multirow}
\usepackage[utf8]{inputenc}
\usepackage{tabularx}
\usepackage[T1]{fontenc}
\usepackage{graphicx}
\usepackage{subfigure}
\usepackage{enumitem}
\usepackage[table,xcdraw]{xcolor}
\usepackage{amsmath}
\usepackage{fontawesome}
\usepackage{makecell}

\usepackage[utf8]{inputenc}

\definecolor{bb}{rgb}{0.12,0.565,1}
\definecolor{gg}{rgb}{0.2,0.8,0.2}
\definecolor{rr}{rgb}{1,0.85,0.2}

\newif\ifdraft
\drafttrue

\hypersetup{
    colorlinks=true,
}

\newcommand{\ours}[0]{Megrez-3B-Omni}
\title{Megrez-Omni Technical Report}

\author{
    Boxun Li$^{1}$\enskip\enskip
    Yadong Li$^{1}$\enskip\enskip
    Zhiyuan Li$^{1}$\enskip\enskip
    Congyi Liu$^{1}$\enskip\enskip
    Weilin Liu$^{1}$\enskip\enskip
    Guowei Niu$^{1}$\enskip\enskip\\
    Zheyue Tan$^{1}$\enskip\enskip
    Haiyang Xu$^{1}$\enskip\enskip
    Zhuyu Yao$^{1}$\enskip\enskip
    Tao Yuan$^{1}$\enskip\enskip
    Dong Zhou$^{1}$\enskip\enskip
    Yueqing Zhuang$^{1}$\enskip\enskip \\
    Shengen Yan$^{1}$\enskip\enskip
    Guohao Dai$^{2}$\enskip\enskip
    Yu Wang$^{3}$\enskip\enskip
    \\
    \textsuperscript{1} Infinigence-AI\thanks{The listing of authors is in alphabetical order based on their last names.} \enskip
    \textsuperscript{2} Shanghai Jiao Tong University  \enskip 
    \textsuperscript{3} Tsinghua University \enskip 
    \enskip \\
}

\begin{document}

\maketitle

\begin{center}
  \vspace{-3em}
  \faGithub~\url{https://github.com/infinigence/Infini-Megrez-Omni}
  \vspace{0.75em}
\end{center}

\begin{figure}[htbp]
  \centering
  \subfigure[]{
        \includegraphics[width=0.44\textwidth]{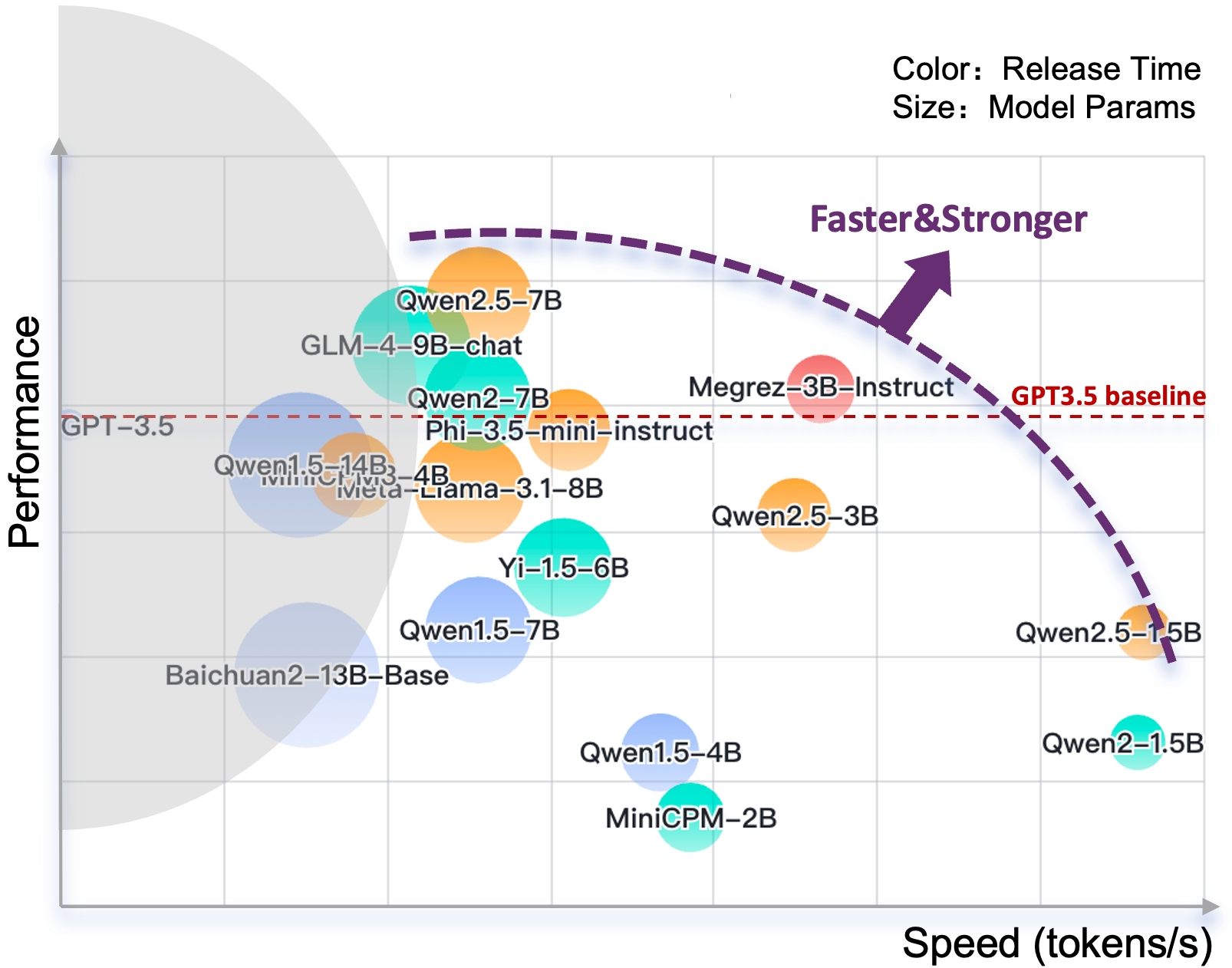}
        \label{Megrez-3B-Instruct}
    }
    \subfigure[]{
        \includegraphics[width=0.51\textwidth]{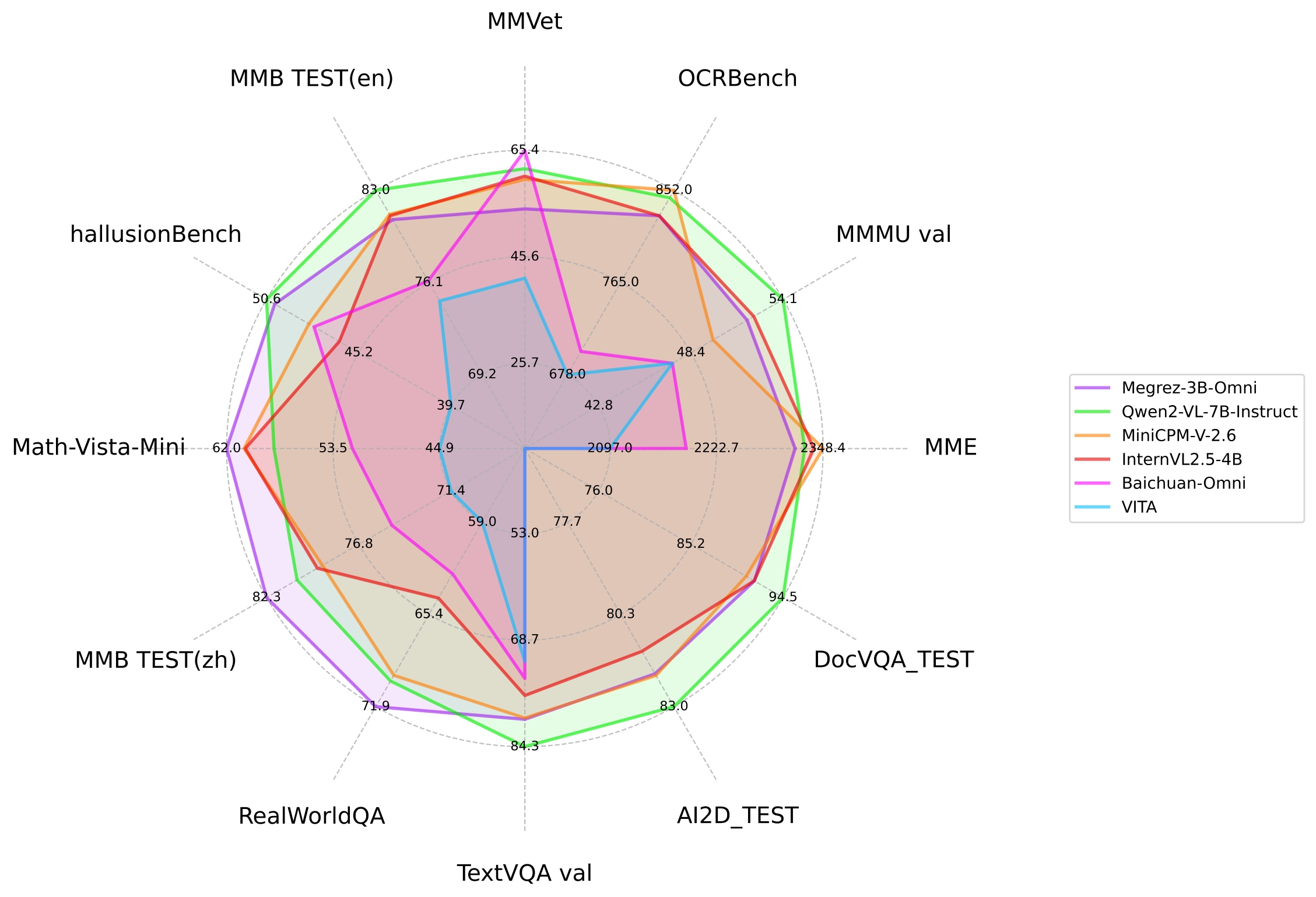}
        \label{Megrez-3B-Omni}
    }
  
  \caption{(a) Despite having the fewest parameters compared to other models, Megrez-3B-Instruct demonstrates superior accuracy on MMLU benchmark. This further extends the capability boundaries of small-scale models, offering new intelligent solutions for edge devices. (b) Megrez-3B-Omni achieves state-of-the-art performance on a broad range of vision tasks compared with other open source models.}
  \label{fig:assets_multitask}
\end{figure}

\begin{abstract}

In this work, we present the Megrez models, comprising a language model (Megrez-3B-Instruct) and a multimodal model (Megrez-3B-Omni). These models are designed to deliver fast inference, compactness, and robust edge-side intelligence through a software-hardware co-design approach. Megrez-3B-Instruct offers several advantages, including high accuracy, high speed, ease of use, and a wide range of applications. 
Building on Megrez-3B-Instruct, Megrez-3B-Omni is an on-device multimodal understanding LLM that supports image, text, and audio analysis. It achieves state-of-the-art accuracy across all three modalities and demonstrates strong versatility and robustness, setting a new benchmark for multimodal AI models.

\end{abstract}

\section{Introduction}

    With the development of large language models (LLMs) and multimodal large language models (MLLMs), there has been significant progress in the model's abilities for understanding, reasoning, and interaction. However, the huge cost of running a LLM or MLLM with massive number of parameters and extensive computation remains a significant challenge.
    Most LLMs and MLLMs are deployed on high-performance cloud servers, which limits their application on edge-side scenarios such as mobile phones, personal computers, vehicles, and robotics.
    
    To adapt to these scenarios, many lightweight LLM models have been proposed. Owing to the progressive enhancement of training corpus quality and the burgeoning availability of distilled data, the performance of relative small models has been significantly augmented~\cite{hu2024minicpm,abdin2024phi3,yang2024qwen2,yang2024qwen2_5}. Edge-side language models can break free from the limitations of remote computing resources and network constraints, allowing users to enjoy the capabilities of language models anytime and anywhere.

    Currently, the development of small language models primarily focuses on understanding and generating purely text-based content~\cite{mit2025techreview}. However, small multimodal models capable of processing text, vision, and speech simultaneously are still significantly lacking. This stands in stark contrast to the application requirements of edge devices. 
    To tackle the need for visual and speech information analysis on edge devices, we specifically designed \ours. The proposed Megrez-3B-Omni is an on-device multimodal understanding LLM model. It is an extension of the Megrez-3B-Instruct model and supports analysis of image, text, and audio modalities. The model achieves state-of-the-art accuracy in all three domains:
    
\begin{itemize}
    \item Language Understanding: Megrez-3B-Omni retains text understanding capabilities without significant trade-offs. Compared to its single-modal counterpart (Megrez-3B-Instruct), the accuracy variation is less than 2\%, maintaining state-of-the-art performance on benchmarks like C-EVAL, MMLU/MMLU Pro, and AlignBench. It also outperforms previous-generation models with 14B parameters.
    
    \item Image Understanding: By utilizing SigLip-400M for constructing image tokens, Megrez-3B-Omni outperforms models with more parameters such as LLaVA-NeXT-Yi-34B. It is one of the best image understanding models among multiple mainstream benchmarks, including MME, MMMU, and OCRBench. It demonstrates excellent performance in tasks such as scene understanding and OCR.
    
    \item  Speech Understanding: Equipped with the encoder head of Qwen2-Audio/whisper-large-v3, the model supports both Chinese and English speech input, multi-turn conversations, and voice-based questions about input images. It can directly respond to voice commands with text and achieved leading results across multiple benchmarks.
\end{itemize}

Through an optimized data processing workflow, strategic selection of training methodologies, and balanced data ratios across modalities, we demonstrate the feasibility of deploying multimodal models at the edge devices. We hope Megrez series can serve as an example for unveiling the potential of edge-side omni models, and help draw more attention to improve the research in this area.

\section{Related Works}

Language models vary in size to suit different scenarios. According to the scaling law~\cite{kaplan2020scaling}, increasing the size of LLMs often enhances their performance in downstream tasks. However, larger models require more computational resources, slower inference speeds, and greater GPU memory. To facilitate the deployment of large models in resource-constrained environments, smaller-sized models like Qwen~\cite{bai2023qwen, chu2023qwen, yang2024qwen2} and MiniCPM~\cite{hu2024minicpm} have been introduced. Qwen, for instance, offers a model with 0.5 billion parameters. To balance resource usage and performance, we propose a 3-billion-parameter language model.

Moreover, the emergent capabilities of large language models (LLMs) have expanded to include the understanding of visual and audio information~\cite{flamingo, chen2022pali, li2023blip, huang2023language, peng2023kosmos, zhu2023minigpt, ye2023mplug, chen2023shikra, zhang2023video, sun2023generative}. Notable models such as CLIP~\cite{clip} and OFA~\cite{wang2022ofa} project visual and textual information into a unified representation space, facilitating downstream multimodal tasks. 
There are two primary methods for integrating vision features into LLMs: aligning vision feature encoders via (i) transformer layers or (ii) multi layer perceptrons. LLaVA~\cite{liu2023improvedllava, liu2024llavanext, liu2023llava} employs a simple projection matrix to connect the pre-trained CLIP ViT-L/14 visual encoder with the Vicuna LLM. Qwen-VLs are a series of high-performance and versatile vision-language foundation models that use transformer layers to link ViT and LLM. In this paper, we deliver the transformer solution for dynamic vision resolution inputs.

For the audio modality, researchers have attempted using well-trained audio foundation models, such as AudioGPT~\cite{huang2024audiogpt} and HuggingGPT~\cite{shen2024hugginggpt}, as tools while employing LLMs as flexible interfaces. These efforts typically involve directing LLMs to generate commands for external tools or converting human speech to text before feeding it into the LLMs. Recent works explored building end-to-end audio-text LLMs for direct speech interaction, such as SpeechGPT~\cite{zhang2023speechgpt}, BLSP~\cite{wang2024blspbootstrappinglanguagespeechpretraining}, and LLaSM~\cite{shu2023llasm}. Furthermore, Qwen-Audio~\cite{chu2023qwen} leveraged the LLaVA architecture, which has been successfully applied in vision-text LLMs, to develop a unified audio-text multi-task multilingual LLMs capable of perceiving and understanding audio inputs while preserving the textual conversational abilities.  

With the development of multimodality, integrating both vision and audio into LLMs has become an attempt to enhance the capabilities of large language models. For example, Vita~\cite{fu2024vita} propose a carefully designed multi-stage training methodology that progressively trains LLM to understand both visual and audio information. Baichuan-omni~\cite{li2024baichuanomni} is an open-source 7B Multimodal Large Language Model (MLLM) which processes and analyzes modalities of image, video, audio, and text. Different from previous work, we introduce Megrez-3B-Omni, an on-device multimodal large language model. Our proposed model has the ability to process visual and audio information, without diminishing the ability of handling text.

\section{Large Language Model}

\subsection{Tokenizer}

In designing tokenizers, it is crucial to balance maintaining an appropriately sized vocabulary for effective word embedding training and achieving a high compression rate for enhanced inference efficiency. Following the design principles of the Qwen2 tokenizer~\cite{yang2024qwen2}, we excluded tokens for languages other than Chinese and English, resulting in a vocabulary size of approximately 120,000 tokens. This reduction from 151,643 to 120,000 tokens aims to balance computational efficiency and model performance. To further improve the compression rate for Chinese, we significantly increased the proportion of Chinese in the training corpus. \autoref{tokenizer_compression_rate} provides a detailed comparison of Megrez's tokenizer with others, demonstrating that our tokenizer achieves the highest compression ratio, particularly in the Chinese corpus.

\begin{table}[h]
\centering
\caption{Compression Rates for Different Models.}
\label{tokenizer_compression_rate}
\begin{tabular}{lcccccc}
\toprule
 & Baichuan2 & ChatGLM2 & Llama2 & MiniCPM & Megrez \\
\midrule
Vocab Size & 125,696 & 64,794 & 32,000 & 122,753 & 120,000 \\
\midrule
\multicolumn{6}{c}{\textbf{Compression Rate (Bytes/Tokens)}} \\
\midrule
Chinese & 3.64 & 3.54 & 1.87 & 3.73 & 5.02 \\
English & 4.12 & 4.02 & 3.78 & 4.14 & 4.28 \\
Code & 2.71 & 2.71 & 2.74 & 2.81 & 2.69 \\
Paper & 2.74 & 2.88 & 2.97 & 2.93 & 3.48 \\
\midrule
Average & 3.30 & 3.29 & 2.84 & 3.40 & 3.86 \\
\bottomrule
\end{tabular}
\end{table}

\subsection{Architecture}

At the beginning of the model design, we debated whether to design a unique but efficient model structure (such as sparse activation and linear attention) or use a widely adopted structure for ease of development. We chose the latter one and adopted the standard LLaMA structure. Our goal is to benefit developers to deploy the model on various platforms without modifications and minimize the complexity of future development.

\subsection{Data}

\subsubsection{Pretrain}

To create an effective pretraining dataset, it is essential that the data be diverse and cover a wide range of types, domains, and tasks. Our dataset is carefully curated to meet these criteria, incorporating public web documents, encyclopedias, books, code samples, and other sources. Additionally, our dataset is multilingual, with a significant portion of the data in English and Chinese.
We implement a comprehensive data preprocessing procedure. We begin by identifying the language using specialized tools. To enhance data diversity, we apply deduplication techniques, such as exact-match deduplication after normalization and fuzzy deduplication using MinHash and LSH algorithms. We then filter out low-quality data through a combination of rule-based and machine-learning-based methods. Specifically, we utilize multiple models to score the content, including language models, text-quality scoring models, and models designed to identify potentially offensive or inappropriate content. Additionally, we manually sample and review texts from various sources to ensure their quality. Finally, we selectively up-sample data from certain sources to ensure that our models are trained on a diverse range of high-quality content.

Inspired by the methods outlined in deepseek-coder~\cite{guo2024deepseek}, we implemented strict filtering and cleaning of GitHub code data. For Python code, we developed more refined readability rules to clean the data, which allowed us to extract high-quality datasets. Based on these cleaned data, we analyzed the reference relationships within individual repositories and applied topological sorting to the code.

\subsubsection{SFT}

Our supervised fine-tuning process is divided into two stages. In the first stage, we perform fine-tuning using a large amount of data with a short context length of 4K. In the second stage, we fine-tune using a mixture of short and long text conversation data with a long context length of 32K.

For our short text data, we utilized open-source instructional datasets that encompass various domains such as dialogue, code, mathematics, instruction following, and complex reasoning. To enhance the complexity of the instructions, we employed a self-evolution method, ensuring that our instructional data spans a diverse range of difficulty levels. 

For our long text data, we curated a selection of high-quality open-source datasets, including tasks such as long text summarization and long text question-answering. To further diversify our long text instructional data, we sourced high-quality long text data. 
Building upon this foundation, we employed LLMs to auto-generate both instructions and responses.

\subsection{Training Stages}

\subsubsection{Pretrain}

We utilized a multi-stage training paradigm, involving different distributions of pretrain data at different stages. Since our base language model was trained from scratch, we aimed to inject and enhance abilities of the model from different perspectives at each stage, which will be detailed as follows.

\textbf{Pretain Stage.} The primary goal of the Pretrain Stage in Megrez is to establish robust language modeling capabilities. At this stage, the majority of the training data is derived from Common Crawl (CC) in both Chinese and English, GitHub code repositories, and a corpus of books. The learning rate was initially set to \(3e{-5}\), then gradually increased to \(3e{-4}\) during the first 3\% of the total training steps in this stage, after which it remained constant until the model was trained on \textasciitilde 2 trillion tokens. Finally, the learning rate was decayed to \(3e{-5}\) following a cosine schedule. During the decay phase, the model was trained on the last part of tokens from high-quality datasets. To optimize training efficiency, the sequence length was set to 4k tokens for this training stage. In anticipation of extending the model's context length to 32k tokens in later stages, the base value in the Rotary Position Embedding (RoPE) framework~\cite{su2024roformer} was set to 5,000,000.

\textbf{Continue Pretrain Stage.} Our objective at this stage was to enhance the model's ability of long-context understanding. To achieve this, we curated a corpus with longer sequence lengths, drawn from books and other high-quality data sources. Since this stage served as an additional training phase following the model's pretraining, a small fraction of the training data from previous stages was also included to prevent any degradation of the model's general capabilities while extending its context length. The initial learning rate for this stage was set to \(3e{-5}\), which decayed to \(3e{-6}\) using a cosine decay schedule. 

\subsubsection{SFT}

\textbf{Megrez-Instruct-4K.} In the first stage, the pre-trained model was fine-tuned using millions of examples with a context length of 4K that cover skills such as instruction following, coding, mathematics, logical reasoning, role-playing, and multilingualism. This fine-tuning was conducted over 3 epochs with a learning rate set to \(1e{-5}\).

\textbf{Megrez-Instruct-32K.} In the second stage, the model was further fine-tuned using mixed instructional data points, which included both short and long text data, with a context length of 32K. This fine-tuning was also conducted over 3 epochs, but with a reduced learning rate of \(5e{-6}\). This stage aimed to improve the model's capability to handle long text data while maintaining its performance on short text data.

By employing this two-stage fine-tuning strategy, we ensured that the model achieved a balanced and robust performance across both short and long text data.

\subsubsection{Web Search}

We have developed a pipeline that enables large language models (LLMs) to utilize web search when answering questions\footnote{https://github.com/infinigence/InfiniWebSearch}. This approach helps reduce hallucinations and improves the quality of the model's responses. Within this pipeline, the LLM can autonomously decide whether to use web search based on the user's query. If web search is employed, the LLM extracts relevant information from the search results and summarizes it. The final answer to the user's question is then formulated based on the summarized information from multiple web sources.

Regarding model training, we introduced a lightweight post-training phase after the supervised fine-tuning (SFT) stage to enhance the model's function-calling capabilities. The training data for this phase includes only a few of the general SFT training data and web search task synthetic data.

\section{Multimodal Large Language Model}

\subsection{Multimodal Model Architecture}

\begin{figure}[htbp]
  \centering
  \includegraphics[width=0.8\textwidth]{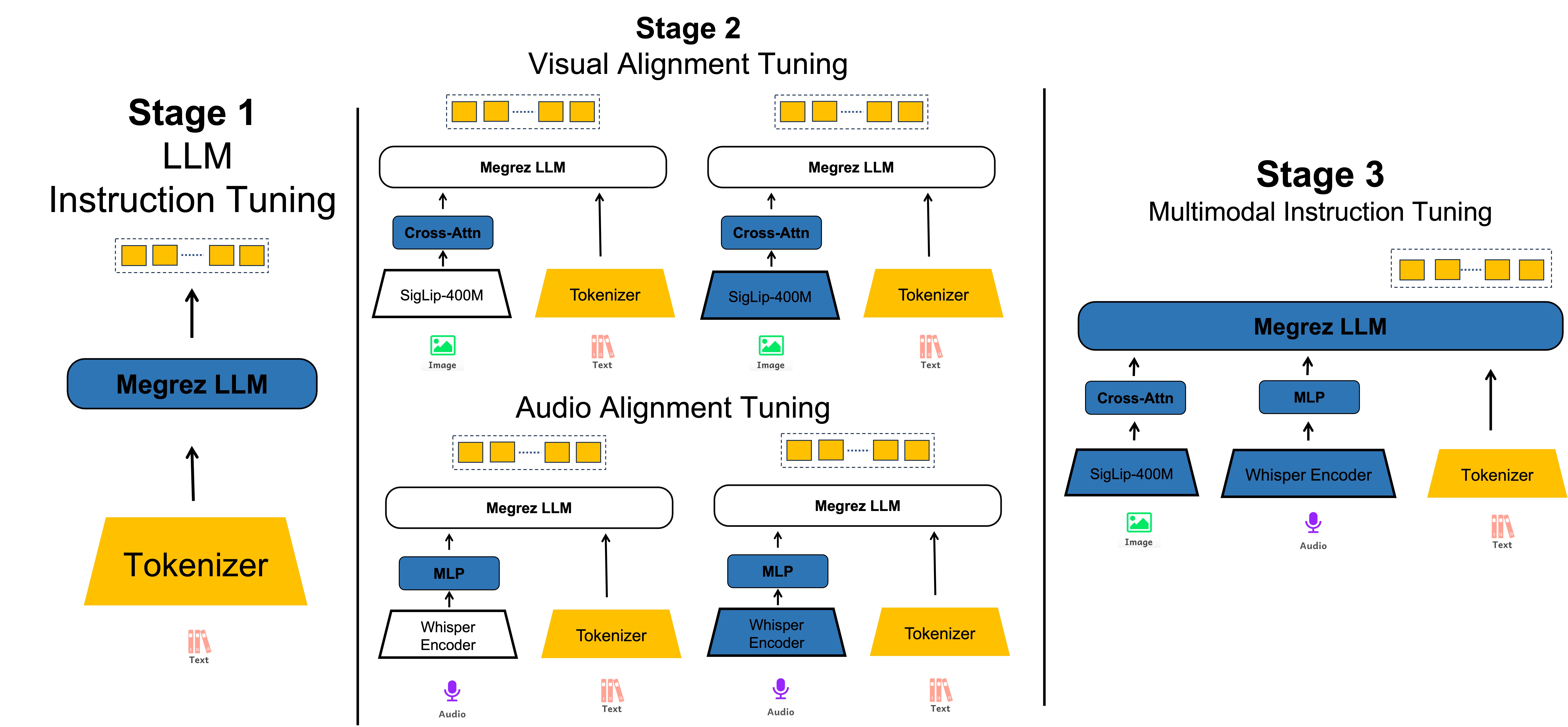}
  \caption{Megrez-O architecture. During the training stage, the white module is frozen and the blue module is trained.}
  \label{fig:architecture}
\end{figure}

   \begin{table}[!ht]
    \centering
    \caption{The parameters count of each module in Megrez-Omni.}
    \label{table:params}
    \begin{tabular}{lccc}
    \toprule
    ~ & Language Module & Vision Module & Audio Module \\
    \midrule
    Architecture & \multirow{2}{*}{Llama-2 with GQA} & \multirow{2}{*}{SigLip-SO400M} & \multirow{2}{*}{Whisper-large-v3} \\
    (encoder-only) & & & \\
    \midrule
    \#  Params & \multirow{2}{*}{2.29B} & \multirow{2}{*}{0.42B} & \multirow{2}{*}{0.64B} \\
    (Backbone) & & & \\
    \midrule
    Connector & - & Cross Attention & Linear \\
    \midrule
    \#  Params & Emb: 0.31B & \multirow{2}{*}{Connector: 0.036B} & \multirow{2}{*}{Connector: 0.003B} \\
    (Others) & Softmax: 0.31B & & \\
    \midrule
    \# Params (Total) & \multicolumn{3}{c}{4B} \\
    \midrule
    \# Vocab Size & 122880 & 64 tokens/slice & - \\
    \midrule
    Context length & \multicolumn{3}{c}{4K tokens} \\
    \midrule
    Supported languages & \multicolumn{3}{c}{Chinese \& English}    \\
    \bottomrule
    \end{tabular}
    \end{table}
     
 The overall architecture of \ours, as shown in \autoref{fig:architecture}, follows the design philosophy of the LLaVA series model, employing a ``Multimodal Encoder-Connector-LLM'' configuration to enhance unified comprehension of image, audio, and language. Megrez-3B-Omni consists of five key components: a visual encoder, a visual compression layer, an audio encoder, an audio compression layer, and an LLM. The input image is processed by the visual encoder using an adaptive encoding method, and the audio input is similarly encoded by the audio encoder. For visual encoding, we use the SigLIP SoViT-400m model, and the generated visual tokens are compressed by a perceiver resampler structure with a single cross-attention layer. For audio, we use the whisper-large-v3 model, trained for speech recognition and translation, and apply a linear transformation to map audio features into the word embedding space. The compressed visual and audio tokens, along with the textual input, are fed into the LLM for conditional text generation.

 Another important setting is in the vision encoder, to address resolution and computational problem in images, we slice original image to keep detail information in vision. For each image slice, the visual output of each section occupies 64 tokens in the LLM. For a single image, we can divide it into a maximum of 9 slices to get local information. In addition, the original image will be resized to around 448$\times$448 to capture global information. A single image can contain a maximum of 9$\times$64 tokens from the slices and 64 tokens from the global image. 

 Regarding the audio encoder, we adopt the architecture of Whisper's encoder and incorporate an MLP projector to align the output features with the embedding space of the LLM. An input audio clip is truncated to a maximum duration of 30 seconds, after which its mel-spectrogram is obtained via a short-time Fourier transform. The output sequence from the Whisper encoder is then compressed to match the dimension of the LLM's embedding space through the MLP projection, producing 50 continuous tokens per second of audio. For raw inputs exceeding 30 seconds, the audio can be segmented into multiple clips and subsequently concatenated within the embedding space to preserve the complete information.

\subsection{Training Data}

\subsubsection{Vision}

The image training data, as outlined in \autoref{tab:vision-train-data}, is composed of diverse categories including caption and question-answering (QA) datasets in both Chinese and English. Throughout various training phases, specific subsets of the dataset are sampled to fulfill distinct objectives. The datasets are organized into the following categories:

\textbf{Image Captioning.} This category includes datasets such as LAION~\cite{schuhmann2022laion}, SA1B~\cite{kirillov2023segany}, ShareGPT4V~\cite{chen2024sharegpt4videoimprovingvideounderstanding}, Objects365-Caption~\cite{shao2019objects365}, etc. These are instrumental in training the model to produce descriptive language for images, enhancing its ability to generate captions that accurately reflect the visual content.

\textbf{Image QA Data.} Comprising datasets like MMInstruct~\cite{liu2024mminstruct}, COCO-QA~\cite{ren2015cocoqa}, VQA~\cite{antol2015vqa}, and ScienceQA~\cite{lu2022scienceqa}, this category focuses on training the model to answer questions based on images and perform visual reasoning tasks. These datasets include both general image QA and science reasoning tasks, which are crucial for developing a model capable of understanding and interpreting visual information in the context of questions.

\textbf{OCR.} This category is designed to support the model's comprehension of Optical Character Recognition (OCR). It utilizes datasets such as SynthText~\cite{gupta2016synthtext}, SynthDoG~\cite{kim2022ocr}, DocVQA~\cite{mathew2021docvqa}, TextVQA~\cite{singh2019textvqa}, and OCR-VQA~\cite{mishra2019ocrvqa}. These datasets are essential for training the model to interpret and understand text within images and diagrams, which is a critical skill for applications involving document analysis and data extraction from visual formats.

By strategically sampling from these datasets during different training stages, we aim to develop a model that is not only capable of generating descriptive language for images but also proficient in answering image-based questions and interpreting visual content. This approach ensures that the model is well-rounded and can handle a variety of tasks related to image understanding and processing.

\begin{table}[!ht]
 \renewcommand\arraystretch{1.3}
    \centering
    \caption{Summary of the pretrain and sft vision-language data mixture of Megrez-3B-Omni.}
    \label{tab:vision-train-data}
    \resizebox{0.97\textwidth}{!}{
    \begin{tabular}{c | l | l }
    \toprule
    Stage & Category &  Sources \\
    \midrule
    \multirow{5}{*}{Pretrain} & \multirow{4}{*}{Captioning} & 
            COCO~\citep{lin2014mscoco}, CC3M~\citep{sharma2018cc3m}, CC12M~\citep{changpinyo2021cc12m}, LAION-COCO~\citep{schuhmann2022laion}, COYO~\citep{kakaobrain2022coyo-700m}, AIC~\citep{wu2017aic},\\      
            & & Laion-EN~\citep{schuhmann2022}, LAION-2B-Chinese~\citep{schuhmann2022laion}, WuKong~\citep{gu2022wukong}, SA-1B-Caption~\citep{kirillov2023segany}, TextCap~\citep{sidorov2020textcaps},\\
            & & MMInstruct~\citep{liu2024mminstruct},
            OpenImages-Caption~\citep{Kuznetsova_2020},
            Objects365-Caption~\citep{shao2019objects365},
            LLaVAR~\citep{zhang2023llavar},
            \\
            & &
            ShareGPT4V~\citep{chen2023sharegpt4v},
            ShareGPT4o~\cite{chen2024far},
            ShareCaptioner~\citep{chen2023sharegpt4v}, Flickr-30K~\citep{plummer2015flickr30k}
            \\
     \cmidrule(lr){2-3} & OCR & 
            WIT~\citep{srinivasan2021wit},  SynthText~\citep{gupta2016synthtext}, SynthDoG-en~\citep{kim2022synthdog}, SynthDoG-zh~\citep{kim2022synthdog}, ArxivCap~\citep{li2024arxivcap}, etc.\\
    \midrule
    \multirow{16}{*}{SFT}& Captioning & sample from Pretrain Captioning data \\
     \cmidrule(lr){2-3} & \multirow{2}{*}{General QA} & 
            VGQA~\citep{krishna2017vg}, IconQA~\citep{lu2021iconqa}, GQA~\citep{hudson2019gqa}, VQAv2~\citep{antol2015vqa} CLEVR~\citep{johnson2017clevr}, VizWiz~\citep{gurari2018vizwiz},\\
            & & Visual7W~\citep{zhu2016visual7w}, COCO-QA~\citep{ren2015cocoqa}, MMInstruct (en\& zh)~\citep{liu2024mminstruct} \\ 
     \cmidrule(lr){2-3} & \multirow{2}{*}{Knowledge} & 
            OKVQA~\citep{marino2019okvqa}, A-OKVQA~\citep{schwenk2022aokvqa}, KVQA~\citep{shah2019kvqa}, ScienceQA~\citep{lu2022scienceqa},ART500K~\citep{mao2017deepart}, \\
            & & KonIQ-10K~\citep{koniq10k}, \textcolor{gray}{Synthetic Knowledge/QA data (zh)} \\
     \cmidrule(lr){2-3} & \multirow{2}{*}{Mathematics} &  
            GeoQA~\citep{chen2021geoqa}, SMART-101~\citep{cherian2023smart101}, MAVIS~\citep{zhang2024mavis}, MetaMath-Rendered~\citep{yu2024metamath}, Geometry3K~\cite{lu2021inter}, \\
            & & \textcolor{gray}{Synthetic arithmetic and geometric mathematical data} \\
     \cmidrule(lr){2-3} & Chart &  
            FSVQA~\citep{shin2016fsvqa}, Visual-Dialog~\citep{das2017visualdialog}, MMTab~\citep{zheng2024multimodaltableunderstanding}, ChartQA~\citep{masry2022chartqa}, MMC-Inst~\cite{liu2024mmc}\\
     \cmidrule(lr){2-3} & \multirow{3}{*}{OCR} & 
            FigureQA~\citep{kahou2017figureqa}, DVQA~\citep{kafle2018dvqa}, DocVQA~\citep{mathew2021docvqa}, TextVQA~\citep{singh2019textvqa}, OCR-VQA~\citep{mishra2019ocrvqa}, ST-VQA~\citep{biten2019stvqa}, \\
            &  & VisualMRC~\citep{tanaka2021visualmrc}, DeepForm~\citep{deepform}, TabFact~\citep{chen2019tabfact}, InfographicsVQA~\citep{mathew2022infographicvqa}, Real-CQA~\citep{ahmed2023realcqa}, \\
            &  & AI2D~\citep{kembhavi2016ai2d}, \textcolor{gray}{Synthetic Handwritten OCR data}, \textcolor{gray}{Synthetic Wikipedia OCR data}  \\
     \cmidrule(lr){2-3} & \multirow{2}{*}{Instruct} & 
            SVIT~\citep{zhao2023svit}, LLaVA-Instruct-150K~\citep{liu2023llava}, UniMM-Chat~\citep{yu2023unimm}, ShareGPT4V~\citep{chen2023sharegpt4v} \\
            &  & LVIS~\citep{gupta2019lvis}, ALLaVA~\citep{chen2024allava}, Cambrain-GPT4o~\citep{tong2024cambrian}, \textcolor{gray}{Synthetic Real-World Instruct data}\\
    \bottomrule
    \end{tabular}}
\end{table}

\subsubsection{Audio}

The audio training data are divided into two categories: pretraining ASR data and audio instruction data, corresponding to the two phases of audio module training. 

Regarding the audio instruction data, we adopted a synthetic approach to ensure efficiency and cost-effectiveness. First, we selected realistic dialogue data from our text SFT data sources to guarantee high-quality conversational content. Text-to-speech (TTS) generation was then applied to all instructions. The TTS model plays a critical role in audio data synthesis, ensuring diversity and naturalness in the generated voices, thus simulating realistic conversational scenarios involving different users. We choose the ChatTTS model due to its fast processing speed, diverse voice styles, and natural tone. In addition, we incorporated human speech recordings to better align with the distribution of realistic applications. To achieve this, we select instruction-like audio clips from the ASR datasets using a combination of rule-based filtering and LLM-based judgement. For these selected audio clips, GPT-4o was employed to generate responses based on their text transcripts. 

\subsection{Training Strategies}

\subsubsection{Vision Alignment}

The alignment process for the vision modality is divided into two distinct stages. Initially, we train the visual connector to establish an initial alignment between image representations and text using an image captioning dataset. During this phase, the visual encoder and the LLM are frozen, while the visual connector is trained with a learning rate of \(3e{-4}\). In the subsequent stage, we freeze the LLM and train both the visual encoder and the connector with a reduced learning rate of \(1e{-4}\). In addition to general VQA tasks, we have specifically synthesized high-quality QA data points for OCR and charts to enhance the model's abstract visual understanding.

\subsubsection{Audio Alignment}
Similar to the vision branch, the alignment of the audio modality also involves two stages. We start to finetune the MLP connector using our ASR dataset while keeping the audio encoder and LLM frozen with a learning rate of \(3e{-4}\). This step aims to align the semantic representations of the audio and text modalities within the embedding space. To maintain consistency in task type and improve generalization in subsequent stages, we reformulate the ASR task into an instruction-following format. Specifically, given an audio clip and a transcription instruction, the model generates the corresponding transcript. Then, we freeze the LLM and train both the visual encoder and the connector with a reduced learning rate of \(3e{-5}\) using audio SFT data. 

\subsubsection{Omni Instruction Tuning}

Since the audio encoder has been pre-trained on extensive audio datasets encompassing ASR, S2TT, AAC, and other audio-related tasks, our model has already achieved satisfactory performance on many speech tasks after the alignment of the audio modality. Nonetheless, a significant semantic gap persists between the visual and textual modalities. Even after the alignment of the visual modality, the model is only capable of performing simple image caption task, and continues to struggle with complex vision tasks, such as OCR, VQA and multi turn conversations.

During the multimodal instruction tuning stage, both the visual encoder and connector, the audio encoder and connector, as well as the LLM are trainable, aiming to improve the visual understanding capabilities. To preserve the capabilities of the text and audio modalities, we employ a smaller learning rate of \(1e{-5}\) to train all parameters with a diverse set of open-source, synthetic, and internally annotated data, across text, audio, image-text, and image-audio modalities.
To ensure the model's seamless transition between various modalities, we randomly shuffle and concatenate data from different modalities to a length of 4K tokens. The data concatenation strategy significantly reduces the training time.

\newcommand{\ouromni}[0]{Megrez-3B-Omni\ }
\newcommand{\ouromnis}[0]{Megrez-3B-Omni's\ }

\section{Experiment}

\subsection{Language Performance}
We conducted a comprehensive evaluation of the model's capabilities, encompassing conversational performance, instruction-following proficiency, general understanding, coding skills, and mathematical reasoning. Specifically, conversational capabilities were assessed using MT-Bench~\cite{zheng2024judging} and AlignBench (ZH)~\cite{liu2023alignbench}, while instruction-following abilities were evaluated through IFEval~\cite{zhou2023instruction}. General understanding was measured using C-EVAL (ZH)~\cite{huang2024c}, CMMLU (ZH)~\cite{li2023cmmlu}, MMLU~\cite{hendrycks2020measuring}, and MMLU-Pro~\cite{wang2406mmlu}. Coding proficiency was tested using HumanEval~\cite{chen2021codex} and MBPP~\cite{austin2021program}, and mathematical reasoning was examined through GSM8K~\cite{cobbe2021gsm8k} and MATH~\cite{hendrycks2021measuring}.

\begin{table}[!ht]
    \centering
    \caption{LLM Evaluation Results}
    \label{tab:llm_eval}
    \resizebox{\textwidth}{!}{%
    \begin{tabular}{@{}l r *{11}{c} @{}}
    \toprule
    \multirow{2}{*}{Model} & Size & \multicolumn{3}{c}{Chat \& Instruction} & \multicolumn{4}{c}{Zh\&En Tasks} & \multicolumn{2}{c}{Code} & \multicolumn{2}{c}{Math} \\ 
    \cmidrule(lr){3-5} \cmidrule(lr){6-9} \cmidrule(lr){10-11} \cmidrule(lr){12-13}
      & {\small(B)} & MT-Bench & AlignBench (ZH) & IFEval & C-EVAL(ZH) & CMMLU(ZH) & MMLU & MMLU-Pro & HumanEval & MBPP & GSM8K & MATH \\
    \midrule
    Qwen1.5-7B & 6.5 & - & - & - & 74.1 & 73.1 & 61.0 & 29.9 & 36.0 & 51.6 & 62.5 & 20.3 \\
    Qwen1.5-7B-Chat & 6.5 & 7.60 & 6.20 & - & 67.3 & - & 59.5 & 29.1 & 46.3 & 48.9 & 60.3 & 23.2 \\
    Qwen1.5-14B & 12.6 & - & - & - & 78.7 & 77.6 & 67.6 & - & 37.8 & 44.0 & 70.1 & 29.2 \\
    Qwen1.5-14B-Chat & 12.6 & 7.90 & - & - & - & - & - & - & - & - & - & - \\
    Qwen2-7B & 6.5 & - & - & - & 83.2 & 83.9 & 70.3 & 40.0 & 51.2 & 65.9 & 79.9 & 44.2 \\
    Qwen2-7B-Instruct & 6.5 & 8.41 & 7.21 & 51.4 & 80.9 & 77.2 & 70.5 & 44.1 & 79.9 & 67.2 & 85.7 & 52.9 \\
    Qwen2.5-3B-Instruct & 2.8 & - & - & - & - & - & - & 43.7 & 74.4 & 72.7 & 86.7 & 65.9 \\
    Qwen2.5-7B & 6.5 & - & - & - & - & - & 74.2 & 45.0 & 57.9 & 74.9 & 85.4 & 49.8 \\
    Qwen2.5-7B-Instruct & 6.5 & 8.75 & - & 74.9 & - & - & - & 56.3 & 84.8 & 79.2 & 91.6 & 75.5 \\
    Llama-3.1-8B & 7.0 & 8.30 & 5.70 & 71.5 & 55.2 & 55.8 & 66.7 & 37.1 & - & - & 84.5 & 51.9 \\
    Llama-3.2-3B & 2.8 & - & - & 77.4 & - & - & 63.4 & - & - & - & 77.7 & 48.0 \\
    Phi-3.5-mini-instruct & 3.6 & 8.60 & 5.70 & 49.4 & 46.1 & 46.9 & 69.0 & 47.4 & 62.8 & 69.6 & 86.2 & 48.5 \\
    MiniCPM3-4B & 3.9 & 8.41 & 6.74 & 68.4 & 73.6 & 73.3 & 67.2 & - & 74.4 & 72.5 & 81.1 & 46.6 \\
    Yi-1.5-6B-Chat & 5.5 & 7.50 & 6.20 & - & 74.2 & 74.7 & 61.0 & - & 64.0 & 70.9 & 78.9 & 40.5 \\
    GLM-4-9B-chat & 8.2 & 8.35 & 7.01 & 64.5 & 75.6 & 71.5 & 72.4 & - & 71.8 & - & 79.6 & 50.6 \\
    Baichuan2-13B-Base & 12.6 & - & 5.25 & - & 58.1 & 62.0 & 59.2 & - & 17.1 & 30.2 & 52.8 & 10.1 \\
    Baichuan-Omni & 7.0 & - & - & - & 68.9 & 72.2 & 65.3 & - & - & - & - & - \\
    VITA & 12.9 & - & - & - & 56.7 & 46.6 & 71.0 & - & - & - & 75.7 & - \\
    \midrule
    \textbf{Megrez-3B-Instruct} & 2.3 & 8.64 & 7.06 & 68.6 & 84.8 & 74.7 & 72.8 & 46.1 & 78.7 & 71.0 & 65.5 & 28.3 \\
    \textbf{Megrez-3B-Omni} & 2.3 & 8.40 & 6.94 & 66.5 & 84.0 & 75.3 & 73.3 & 45.2 & 72.6 & 60.6 & 63.8 & 27.3 \\
    \bottomrule
    \end{tabular}}
\end{table}

The results demonstrate that Megrez-3B-Instruct exhibits superior performance in conversational capabilities, instruction-following abilities, and comprehensive understanding, surpassing other models of similar parameter scales, such as Qwen2.5-3B-Instruct and MiniCPM3-4B. However, its performance in coding and mathematical reasoning remains less competitive.

Additionally, Megrez-3B-Omni shows notable proficiency as a general-purpose large language model (LLM), holding its ground even in comparisons with larger models exceeding 7B parameters, such as Baichuan-Omni and VITA.

\subsection{Vision Performance}

\begin{figure}[htbp]
  \centering
  \includegraphics[width=0.9\textwidth]{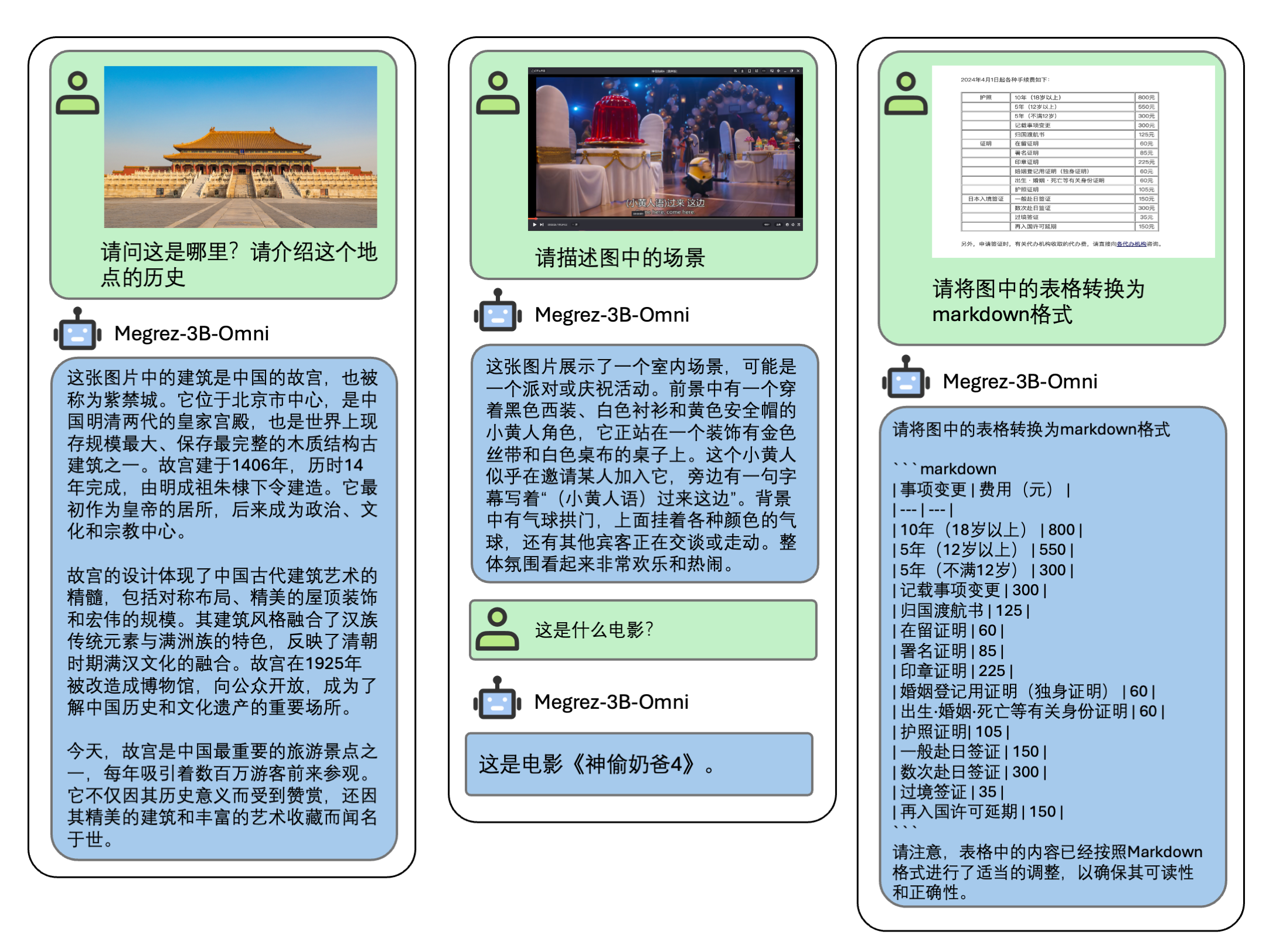}
  \caption{Qualitative results of Megrez-3B-Omni in captioning, reading chart in images and general chat.}
  \label{fig:case4}
\end{figure}

\begin{figure}[htbp]
  \centering
  \includegraphics[width=0.9\textwidth]{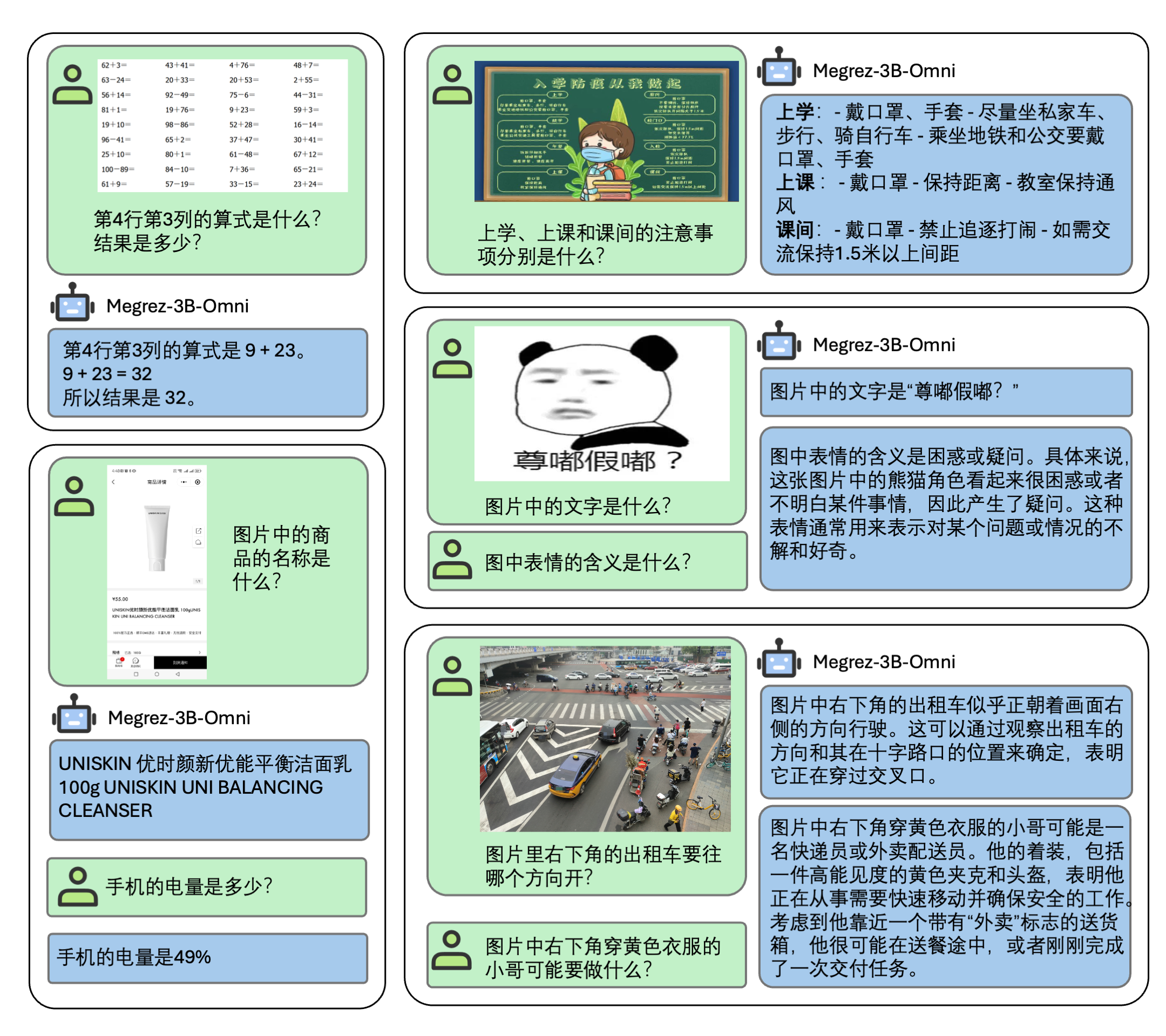}
  \caption{Qualitative results of Megrez-3B-Omni in math, OCR, GUI information, conversation and Reasoning.}
  \label{fig:case5}
\end{figure}

  To rigorously evaluate \ouromnis performance on vision tasks, we employ a diverse suite of benchmark datasets. These include established, classic datasets as well as novel benchmarks provided by VLMEvalKit~\citep{duan2024vlmevalkit}. This comprehensive selection encompasses a wide range of categories, ensuring a thorough and balanced assessment of \ouromnis capabilities across various vision-related tasks. A more intuitive understanding of Megrez-3B-Omni capabilities on multiple vision tasks is shown in  \autoref{fig:case4} and \autoref{fig:case5}.

\textbf{Benchmarks.} \quad 
Refer to OpenCompass~\citep{2023opencompass}, We evaluate \ouromnis performance across 12 representative vision-language benchmarks as follows:
\begin{itemize}
    \item \textbf{General Benchmarks.} Our evaluation includes a comprehensive evaluation of visual question answering, multimodal conversation, knowledge, reasoning and spatial understanding capabilities. Specifically, we employ the following benchmarks: MME~\citep{fu2024mme}, MMBench-EN, MMBench-CN~\citep{liu2023mmbench}, MMMU~\citep{yue2023mmmu}, MMVet~\citep{yu2023mm}, MathVista~\citep{lu2023mathvista}, MStar~\citep{chen2024we}, and RealWorldQA~\citep{grok15}.
    \item \textbf{OCR Benchmarks.} To evaluate OCR capabilities, we utilize four established benchmark, including OCRBench~\citep{liu2024ocrbench}, TextVQA~\citep{singh2019textvqa}, DocVQA~\citep{mathew2021docvqa}, and AI2D~\citep{kembhavi2016diagram}.
    \item \textbf{Hallucination Benchmark.} We incorporate HallusionBench~\citep{guan2024hallusionbench} to assess the trustworthiness and factual correctness of the model's responses.
\end{itemize}

\begin{table}[!ht]
    \centering
    \caption{Experimental results on general vision-language benchmarks. RW QA: RealWorldQA. turbo: GPT-4-Turbo. $\dagger$: The official report lacks precise information regarding the model's total parameter. The best open-source results of models below 10B parameters are highlighted in \textbf{bold}.}
    \label{tab:mllm_general_comparison}
    \resizebox{0.97\textwidth}{!}{
    \begin{tabular}{lccccccccc}
        \toprule
        \multirow{2}{*}{Model} & {Size} & {Open-} & {MME} & {MMB Test} & {MMMU} & {MMVet} & {Math} & {MMstar} & {RW} \\
        & {\small(B)} & {Compass} & {\small (sum)} & {\small(EN/CN)} & {\small(val)} & {\small(turbo)} & {Vista}& & {QA} \\
        \midrule
        Qwen2-VL-2B-Instruct & 2.21 & 57.2 & 1872 & {74.9 / 73.5} & 41.1 & 49.5 & 43 & 48 & 62.9 \\
        InternVL2.5-2B & 2.21 & 59.9 & 2138 & {74.7 / 71.9} & 43.6 & 60.8 & 51.3 & 53.7 & 60.1 \\
        BlueLM-V-3B & 3.1 & 66.1 & {-} & {\textbf{83.0} / 80.5} & 45.1 & 61.8 & 60.8 & \textbf{62.3} & 66.7 \\
        InternVL2.5-4B & 3.71 & 65.1 & 2337 & {81.1 / 79.3} & 52.3 & 60.6 & 60.5 & 58.3 & 64.3 \\
        Baichuan-Omni & 7$^\dagger$ & {-} & 2186 & {76.2 / 74.9} & 47.3 & \textbf{65.4} & 51.9 & {-} & 62.6 \\
        MiniCPM-V-2.6 & 8.1 & 65.2 & 2348 & {81.2 / 79.0} & 49.8 & 60 & 60.6 & 57.3 & 69.7 \\
        Qwen2-VL-7B-Instruct & 8.29 & \textbf{67} & \textbf{2326} & {\textbf{83.0} / 80.5} & \textbf{54.1} & 62 & 58.2 & 60.7 & 70.1 \\
        MiniCPM-Llama3-V-2.5 & 8.54 & 58.8 & 2024 & {77.2 / 74.2} & 45.8 & 52.8 & 54.3 & {-} & 63.5 \\
        \midrule
        VITA & 8$\times$7 & {-} & 2097 & {74.7 / 71.4} & 47.3 & 41.6 & 44.9 & {-} & 59 \\
        GLM-4V-9B & 13.9 & 59.1 & 2018 & {81.1 / 79.4} & 46.9 & 58 & 51.1 & 58.7 & {-} \\
        LLaVA-NeXT-Yi-34B & 34 & 55 & 2006 & {81.1 / 79.0} & 48.8 & 50.7 & 40.4 & 51.6 & 66 \\
        Qwen2-VL-72B-Instruct & 73.4 & 74.8 & 2482 & {86.5 / 86.6} & 64.5 & 74 & 70.5 & 68.3 & 77.8 \\
        \midrule
        Megrez-3B-Omni & 3.38 & 66.2$^\star$ & 2315 & {80.8 / \textbf{82.3}} & 51.9 & 60 & \textbf{62} & 60.5 & \textbf{71.9} \\
        \bottomrule
    \end{tabular}}
\end{table}

We compare with strong baselines in different series, including Qwen2-VL-2B-Instruct~\citep{Qwen2VL}, InternVL2.5-2B~\citep{chen2024expanding}, BlueLM-V-3B~\citep{lu2024bluelmv3b}, InternVL2.5-4B~\citep{chen2024expanding}, Baichuan-Omni~\citep{li2024baichuanomni}, MiniCPM-V-2.6~\citep{yao2024minicpmv}, Qwen2-VL-7B-Instruct~\citep{Qwen2VL}, MiniCPM-Llama3-V-2.5~\citep{yao2024minicpmv}, VITA~\citep{fu2024vita}, GLM-4V-9B~\citep{glm2024chatglm}, LLaVA-NeXT-Yi-34B~\citep{ai2024yi} and Qwen2-VL-72B-Instruct~\citep{Qwen2VL}.

\textbf{Results on General vision-language Benchmarks.} \quad 
As shown in Table~\ref{tab:mllm_general_comparison}, our analysis reveals several key observations. Firstly, \ouromni demonstrates superior performance compared to strong open-source models of comparable size. Secondly, it achieves competitive results with significantly larger models (below 10B parameters), such as MiniCPM-V-2.6 and Qwen2-VL-7B-Instruct, even achieving state-of-the-art performance on MMBench-CN, MathVista, and RealWorldQA benchmarks. Finally, compared to prominent open-source model in the industry, such as Qwen2-VL-72B-Instruct, \ouromni achieves over 80\% accuracy with a parameter count that is an order of magnitude smaller. In summary, these findings indicate that \ouromni achieves a favorable balance between performance and efficiency, making it suitable for broader community adoption and diverse applications.

\begin{table}[!ht]
    \centering
    \caption{Experimental results on OCR and Hallucination benchmarks. $\dagger$: The official report lacks precise information regarding the model's total parameter. The best open-source results of models below 4B parameters are highlighted in \textbf{bold}.}
    \label{tab:mllm_other_comparison}
    \resizebox{0.97\textwidth}{!}{
    \begin{tabular}{lccccccccc}
        \toprule
        \multirow{2}{*}{Model} & {Size} & {OpenCompass} & {OCRBench} & {TextVQA} & {DocVQA} & {AI2D} & {HallusionBench} \\
        & {\small(B)} &  &  & {\small(val)} & {\small(test)} & {\small(test)} & {\small(avg)} \\
        \midrule
        Qwen2-VL-2B-Instruct & 2.21 & 57.2 & 79.4 & 79.7 & 90.1 & 74.7 & 41.7 \\
        InternVL2.5-2B & 2.21 & 59.9 & 80.4 & 74.3 & 88.7 & 74.9 & 42.6 \\
        BlueLM-V-3B & 3.1 & 66.1 & \textbf{82.9} & 78.4 & 87.8 & \textbf{85.3} & 48 \\
        InternVL2.5-4B & 3.71 & 65.1 & 82.8 & 76.8 & \textbf{91.6} & 81.4 & 46.3 \\
        \midrule
        Baichuan-Omni & 7$^\dagger$ & {-} & 70 & 74.3 & {-} & {-} & 47.8 \\
        MiniCPM-V-2.6 & 8.1 & 65.2 & 85.2 & 80.1 & 90.8 & 82.1 & 48.1 \\
        Qwen2-VL-7B-Instruct & 8.29 & 67 & 84.5 & 84.3 & 94.5 & 83 & 50.6 \\
       MiniCPM-Llama3-V-2.5 & 8.54 & 58.8 & 72.5 & 76.6 & 84.8 & 78.4 & 42.4 \\
        VITA & 8$\times$7 & 67.8 & 71.8 & {-} & {-} & 39.7 \\
        GLM-4V-9B & 13.9 & 59.1 & 77.6 & {-} & {-} & 81.1 & 46.6 \\
        LLaVA-NeXT-Yi-34B & 34 & 55 & 57.4 & 69.3 & {-} & 78.9 & 34.8 \\
        Qwen2-VL-72B-Instruct & 73.4 & 74.8 & 87.7 & 85.5 & 96.5 & 88.1 & 58.1 \\
        \midrule
       Megrez-3B-Omni & 3.38 & 66.2 & 82.8 & \textbf{80.3} & \textbf{91.6} & 82.1 & \textbf{50.1} \\
        \bottomrule
    \end{tabular}}
\end{table} 

\textbf{Results on OCR and Hallucination Benchmarks.} \quad 
The experimental results presented in Table~\ref{tab:mllm_other_comparison} indicate that \ouromni demonstrates robust OCR and hallucination mitigation capabilities at comparable parameter sizes. This includes understanding of scene text, documents, chart and screenshots.

\begin{figure}[htbp]
  \centering
  \includegraphics[width=0.8\textwidth]{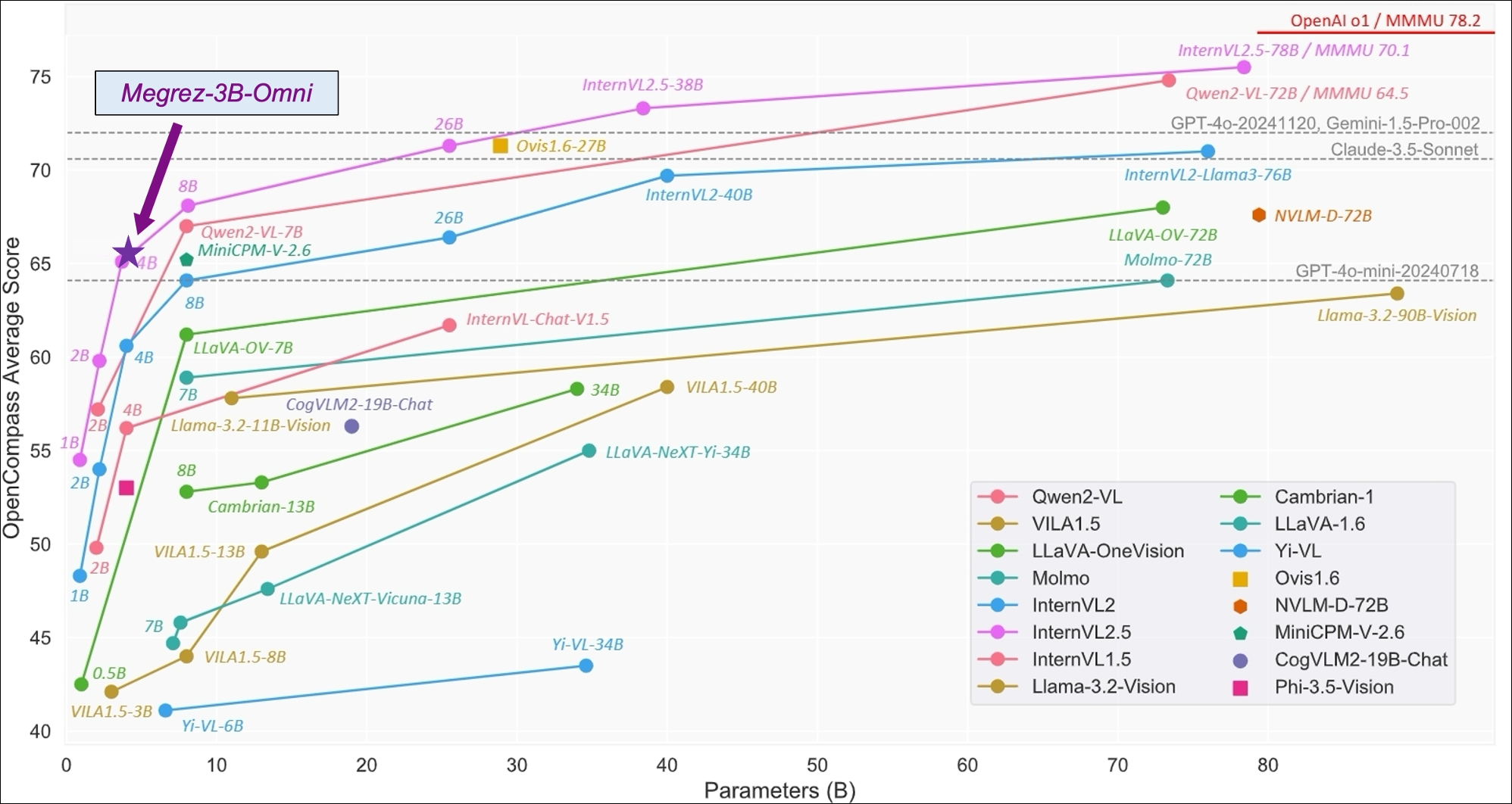}
  \caption{The performance of Megrez-3B-Omni on the OpenCompass test set.}
  \label{fig:assets_opencompass}
\end{figure}

\subsection{Audio Performance}

\begin{table}[t]
\centering
\caption{ASR performence}
\label{tab:asr_perf}
\begin{tabular}{c|cccc}
\toprule
\multirow{2}{*}{Model} & \multirow{2}{*}{Base Model} & Fleurs & WenetSpeech \\
 & & test-zh & test-meeting \\

\midrule
Whisper-large-v3 & -   & 12.4 &  30.8 \\
Qwen2-Audio-7B   & Qwen2-7B  & 9 &  10.7 \\
Baichuan2-omni   & Unknown-7B & 7  &  8.4 \\
VITA             & Mixtral 8x7B & -  & 16.5 \\
\midrule
Megrez-3B-Omni   & Megrez-3B-Instruct  & 10.8  & 16.4 \\
\bottomrule
\end{tabular}
\end{table}

We evaluated the Chinese ASR performance of Megrez-3B-O on the Fleur-test-zh and WenetSpeech-test-meeting datasets. The comparison of results with other multi-modal large language models (LLMs) is presented in \autoref{tab:asr_perf}. Benefiting from targeted training on Chinese ASR datasets, our model achieves superior performance compared to Whisper-large-v3. However, when compared to Qwen2-audio, our model exhibits slightly limited performance due to the disparity in dataset size.

\subsection{Speed Performance}

We conducted speed tests on various language models using the NVIDIA A100 GPU. To ensure fair comparison, all models were deployed using the vLLM inference framework. Both the input and output token counts were set to 128, and the batch size was set to 8\footnote{For more details please refer to:  https://huggingface.co/Infinigence/Megrez-3B-Instruct/blob/main/README\_SPEED.md}. As shown in \autoref{Megrez-3B-Instruct}, our model achieves the fastest inference speed among models with a parameter size around 3 billion.

\section{Conclusion}

In this work, we have open-sourced Megrez-3B-Omni as a step toward developing a truly omni-modal LLM that integrates all human senses. Our Megrez-3B-Omni has achieved leading levels in integrating comprehension across image, text, and audio. 
Despite its promising performance, there remains significant room for improvement in the foundational capabilities across each individual modality. This includes (1) supporting multi-image and video understanding; (2) developing an end-to-end TTS system integrated with LLMs; (3) enhancing perception capabilities for complex visual tasks; (4) supporting real-time video and audio streams in live interactions; and (5) solving complex reasoning problems with reinforcement learning.

\bibliography{references}
\bibliographystyle{plain}

\newpage
\appendix

\end{document}